\documentclass[10pt,twocolumn,letterpaper]{article}
\usepackage[pagenumbers]{wacv}
\usepackage{times}
\usepackage{graphicx}
\usepackage{amsmath,amssymb}
\usepackage{booktabs}
\usepackage{multirow}
\usepackage{url}
\definecolor{wacvblue}{rgb}{0.21,0.49,0.74}
\usepackage[breaklinks,colorlinks,allcolors=wacvblue]{hyperref}

\def\wacvPaperID{1707}
\def\confName{WACV}
\def\confYear{2027}

\graphicspath{{figs/}}

\begin{document}

\title{When Does Corruption Break the Plan? \\ A Degradation-Tolerance Benchmark for Camera-Only End-to-End Driving}
\author{Haohua Que \qquad Handong Yao\thanks{Corresponding author: Handong Yao (handong.yao@uga.edu).}\\
College of Engineering, University of Georgia, Athens, Georgia, USA\\
{\tt\small hq10606@uga.edu \qquad handong.yao@uga.edu}}
\maketitle

\begin{abstract}
Camera-only end-to-end (E2E) driving models are nearing deployment, where the camera stream is degraded
by blur, noise, low light, weather, frame loss, and memory faults. How much a policy tolerates before
its driving breaks is unclear. Corruption-robustness benchmarks target detection or bird's-eye-view
perception, not the planning output that drives the car. We present \textbf{DriveDegrade}, a benchmark
for image-degradation tolerance in camera-only E2E driving. Sixteen corruption families at five
severities are injected on the fly inside the image loader, one operator reaching fifteen policies, and
we evaluate open-loop planning on nuScenes and NAVSIM plus a CARLA closed-loop anchor. First, mild
degradation barely affects planning, and the families that break it have a clear threshold at mid
severity. Second,
fragility is corruption-dependent: blur, JPEG, and raindrop damage planning most, while weather and bit
error are tolerated far into the range. Third, a flat curve is ambiguous, so we separate corruptions
that degrade the image from those that remove it. A planner that reads its camera must lose accuracy
when information is deleted, whatever it does under quality loss. On these two axes the planners
separate sharply, quantifying the ego-status shortcut without mistaking indifference for robustness. A
released vision-language-action planner is flat on both axes, and blinding all six of its cameras costs
it only $11.5\%$.
\end{abstract}

\begin{figure*}[t]\centering
  \includegraphics[width=\textwidth]{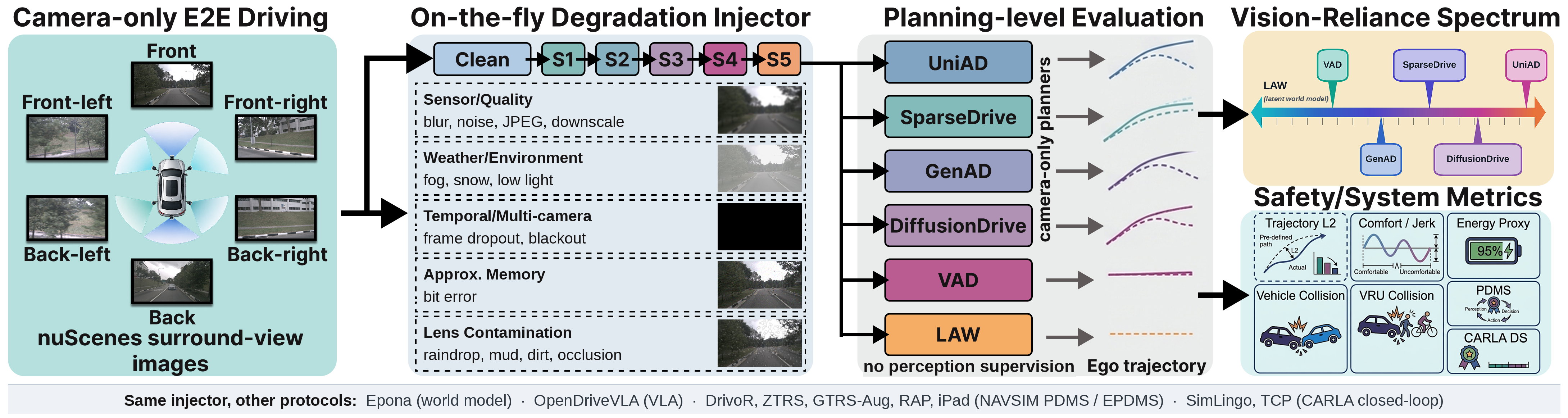}
  \caption{\textbf{DriveDegrade overview.} Surround-view camera streams are degraded on the fly inside the image
  loader over a five-level severity ladder. Six camera-input E2E planners are evaluated at the planning
  level, then summarized through tolerance curves, a vision-reliance diagnostic separating quality
  degradation from information removal, and safety/system metrics. The band beneath lists the nine
  further policies the identical injector reaches under their own protocols. All sixteen corruption
  families are specified in Appendix~\ref{supp:excluded}.}
  \label{fig:overview}
\end{figure*}

\section{Introduction}
\label{sec:intro}
End-to-end (E2E) camera-only driving has matured quickly: a single network reads the surround-view
cameras and predicts the ego trajectory, folding perception, prediction, and planning into one
differentiable stack, as in UniAD~\cite{uniad} and VAD~\cite{vad}. Dropping LiDAR makes these systems
cheaper to deploy, and recent models such as SparseDrive~\cite{sparsedrive}, GenAD~\cite{genad}, and
DiffusionDrive~\cite{diffusiondrive} rival sensor-rich pipelines on open-loop nuScenes~\cite{nuscenes}.
Camera-only E2E driving is no longer a research curiosity but a serious candidate for real vehicles.

A real camera stream is rarely clean. Lenses defocus and the vehicle vibrates; sensors add noise in low
light; weather scatters and occludes the scene; compression and transmission cost quality; frames
arrive late or not at all; and image buffers run at reduced refresh to save power flip bits. Each is
routine in deployment, and each changes the very pixels the planner consumes. How much of this can an
E2E policy absorb before its driving breaks, and which kinds matter most? The answer tells a system
designer where to spend the sensing and compute budget, which failures are merely uncomfortable and
which are dangerous, and how far the imaging pipeline can be compressed or down-clocked before safety
erodes.
Existing corruption benchmarks stop one stage too early: they corrupt the input and then measure
detection or bird's-eye-view perception~\cite{robobev,dong3dcorr,imagenetc}. Perception is not what
drives the car. A planner can stay robust where perception is fragile, or fail where perception still
looks healthy, because it can lean on ego motion and the navigation command rather than the
image~\cite{bevplanner}.

We answer with DriveDegrade (Fig.~\ref{fig:overview}), which injects a recognized corruption taxonomy at
five severities, aligned with ImageNet-C and nuScenes-C~\cite{imagenetc,robobev} and extended with four
driving-specific contamination families, directly into each model's image loader, so the same operator
at the same severity reaches every architecture and no corrupted copies touch disk. We evaluate
open-loop planning on nuScenes with trajectory L2, collision, comfort, and an energy proxy. We repeat a
slice on NAVSIM~\cite{navsim,navsimv2} under its rule-based PDM score (PDMS) and extended PDM score
(EPDMS), so the findings do not rest on imitation error alone. We anchor the trends closed-loop with two
camera-only agents, SimLingo~\cite{simlingo} and TCP~\cite{tcp}, in CARLA~\cite{carla} across three
towns.

\noindent\textbf{Contributions.}
\begin{itemize}\itemsep0pt
  \item \textbf{One injector, fifteen policies, five architecture families.} Planning-level corruption
        evaluation is not new~\cite{robuste2e}; what is new is reaching every architecture with the
        \emph{same} operator at the \emph{same} severity, from a model-agnostic injector inside each
        model's own loader (Sec.~\ref{sec:bench},~\ref{sec:navsim4},~\ref{sec:newmodels},~\ref{sec:carla}).
  \item \textbf{Per-corruption tolerance thresholds.} An analysis that locates, per corruption, the
        severity at which planning starts to break, over sixteen families including four
        driving-specific contamination families whose damage is spatially concentrated rather than
        global (Sec.~\ref{sec:results},~\ref{sec:driving}).
  \item \textbf{A vision-reliance diagnostic that resolves the flat-curve ambiguity.} Separating
        quality degradation from information removal turns the degradation curve into an
        interpretable measure of camera reliance, rather than a robustness score that rewards a
        planner for ignoring its input (Sec.~\ref{sec:cross}).
\end{itemize}

\section{Related Work}
\label{sec:related}
\noindent\textbf{Corruption robustness and degraded-driving benchmarks.}
Common-corruption robustness began with ImageNet-C~\cite{imagenetc} and has since spanned distribution
shift, shape bias, augmentation, backbones, segmentation, and adverse
weather~\cite{objectnet,imageneta,stylizedimagenet,augmix,aresbench,kamannseg,michaelis,acdc,foggycityscapes},
then driving perception in three dimensions and
LiDAR~\cite{robobev,dong3dcorr,robo3d,cc3d,robodrive}. All share one boundary: they report how much
\emph{perception} degrades, not the planned trajectory that moves the vehicle.

Two works probe E2E driving robustness directly. RobustE2E~\cite{robuste2e} pairs five adversarial
attacks and a module-wise attack with 17 natural corruption types at five severities on
UniAD~\cite{uniad}, ST-P3~\cite{stp3}, and VAD~\cite{vad}, plus closed-loop case studies. Bench2Drive-Robust~\cite{bench2driverobust}
evaluates four closed-loop agents in CARLA~\cite{carla}, namely TCP~\cite{tcp}, UniAD, VAD, and
SimLingo~\cite{simlingo}, under deployment perturbations: burst frame drop, black-rectangle partial
observation, GPS and speed noise, and inference latency. It targets system-level rather than image-level
imperfection. Both report results across their severity settings.

DriveDegrade differs from both in three ways. \emph{Coverage:} one injector inside each model's own
loader reaches fifteen policies, including the sparse, generative, diffusion, world-model, and
vision-language-action (VLA) planners that the RobustE2E study predates, and a new planner needs one
loader hook (Appendix~\ref{supp:protocol}). \emph{Diagnosis:} neither prior work separates quality
degradation from information removal, so neither can tell a robust curve from an unread camera
(Sec.~\ref{sec:cross}). \emph{Resolution:} neither reports per-family knee locations, driving-specific
lens contamination, an approximate-memory fault family, or a memory-interface energy proxy. DriveDegrade
is complementary to Bench2Drive-Robust in particular: it measures the image-degradation axis that
benchmark excludes by design, and its occlusion family matches that benchmark's partial-observation
perturbation, so the two ladders meet at one point (Sec.~\ref{sec:taxonomy}).

\noindent\textbf{Camera-only end-to-end driving and ego-status shortcuts.}
A growing family predicts the ego plan directly from cameras: UniAD couples detection, tracking,
mapping, motion, and occupancy before planning~\cite{uniad}; VAD and VADv2 use a vectorized
scene~\cite{vad,vadv2}; SparseDrive is sparse query-based~\cite{sparsedrive}; GenAD and DiffusionDrive
cast planning as generation~\cite{genad,diffusiondrive}; LAW plans from a learned latent~\cite{law}.
Others round out the space, some adding
LiDAR~\cite{paradrive,hydramdp,goalflow,stp3,transfuser,transfuserpp,drivetransformer,occnet,fusionad,drivelm},
all building on camera BEV
backbones~\cite{bevformer,bevdet,bevdepth,petr,petrv2,detr3d,lss,fcos3d,bevfusion,solofusion,sparse4d,cam4docc}.
They differ sharply in how much of the plan depends on the image, yet they are normally compared only by
clean accuracy. Open-loop nuScenes planning is largely predictable from ego velocity, acceleration, and
the high-level command, so a model can score well while barely using its
cameras~\cite{bevplanner,admlp}. This is usually treated as a caveat; we treat it as a measurement,
provided the probe distinguishes tolerating a degraded image from disregarding it
(Sec.~\ref{sec:cross}).

\noindent\textbf{Planning evaluation protocols, robustness methods, and energy.}
Open-loop planning is measured on nuScenes and related
datasets~\cite{nuscenes,waymo,kitti,argoverse,argoverse2,bdd100k}, closed-loop behavior in simulators
and planning benchmarks~\cite{nuplan,carla,bench2drive,navsim,safety2drive}. Neither includes systematic
camera-degradation evaluation, and no prior planning-level study spans the modern nuScenes planners at
per-family resolution (Table~\ref{tab:cmp}). Orthogonal lines improve
robustness at training or test time~\cite{fgsm,pgd,rp2,ttt,tent} or study domain
shift~\cite{dgsurvey,shift,seeinthedark,dodgekaram}; we instead \emph{measure} tolerance. Our energy
proxy is grounded in bus-encoding and approximate-memory
work~\cite{businvert,rahaddram,eden,stutzbiterror,rtlnn}.

\begin{table}[t]\centering\footnotesize
\caption{DriveDegrade versus existing robustness and evaluation benchmarks. Perception benchmarks stop at
perception; planning benchmarks do not corrupt the image. RobustE2E~\cite{robuste2e} and
Bench2Drive-Robust~\cite{bench2driverobust} evaluate planning under perturbation and both report graded
severities; Sec.~\ref{sec:related} states their scope. \emph{eval.\ level} is the scored output (BEV,
bird's-eye view; OL, open loop; CL, closed loop). \emph{img.\ corr.} is whether the image itself is
degraded (adv., adversarial; sys., system-level; partial, only some corruptions act on the image).
\emph{policies} counts evaluated E2E policies, ``many'' where a benchmark hosts an open leaderboard.
Ours are six nuScenes planners, a world model, a vision-language-action (VLA) planner, five NAVSIM
policies, and two CARLA agents; a camera+LiDAR reference and an image-blind floor are uncounted
controls. \emph{quality vs.\ removal} is whether a benchmark separates quality degradation from
information removal (Sec.~\ref{sec:cross}).}
\label{tab:cmp}
\setlength{\tabcolsep}{2pt}
\begin{tabular*}{\linewidth}{@{\extracolsep{\fill}}lcccc@{}}
\toprule
benchmark & eval.\ level & img.\ corr. & policies &
  \begin{tabular}[b]{@{}c@{}}quality vs.\\ removal\end{tabular} \\
\midrule
ImageNet-C~\cite{imagenetc}        & class.          & yes        & --   & no \\
Dong et al.~\cite{dong3dcorr}      & 3D det.         & yes        & --   & no \\
RoboBEV~\cite{robobev}             & BEV perc.       & yes        & --   & no \\
Robo3D~\cite{robo3d}               & 3D det./seg.    & LiDAR      & --   & no \\
RoboDrive~\cite{robodrive}         & BEV/occ./depth  & yes        & --   & no \\
Bench2Drive~\cite{bench2drive}     & plan.\,(CL)     & no         & many & no \\
NAVSIM~\cite{navsim}               & plan.\,(OL)     & no         & many & no \\
Safety2Drive~\cite{safety2drive}   & plan.\,(CL)     & partial    & --   & no \\
RobustE2E~\cite{robuste2e}         & plan.\,(OL+CL) & yes+adv. & 3 & no \\
B2D-Robust~\cite{bench2driverobust} & plan.\,(CL) & sys.-level & 4 & no \\
\textbf{Ours}                      & \textbf{plan.\,(OL+CL)} & \textbf{yes} & \textbf{15} & \textbf{yes} \\
\bottomrule
\end{tabular*}
\end{table}

\section{The DriveDegrade Benchmark}
\label{sec:bench}

\subsection{On-the-fly corruption injection}
\label{sec:inject}\label{sec:seedcheck}
Corruptions are applied inside the image loader, after decoding and before normalization, so no
corrupted copies are written to disk and every model consumes the same operator at the same severity
parameter. The injector depends only on numpy/cv2, which is what lets one implementation serve every
model environment and both simulators unchanged.

\noindent\textbf{Scope of identical injection.} Six families are deterministic functions of the input
frame, so every model receives bit-identical pixels. The rest draw a per-frame random realization, and
for those what is shared is the operator and its severity. The released injector seeds each draw as
$\mathrm{blake2b}(\text{seed},\text{frame},\text{camera})$. The realization itself does not matter at
this scale: each cell averages $6019$ frames $\times$ six cameras, and three fresh seeds shift
retention by $0.004$ and $0.002$ on the two most stochastic cells, against per-family effects of
$0.1$--$0.5$ (Appendix~\ref{supp:seeding}).

\subsection{Corruption taxonomy and severity specification}
\label{sec:taxonomy}
The taxonomy follows ImageNet-C and nuScenes-C/RoboBEV across four groups: sensor/quality,
weather/environmental, temporal/multi-camera, and approximate-memory. To these we add a fifth group
specific to a camera carried on a vehicle rather than inherited from generic image robustness,
\emph{lens and windshield contamination}, comprising occlusion, mud splash, raindrop, and lens dirt.
These four differ from blur or noise in kind rather than degree. Their damage is spatially
concentrated, so part of the frame stays pristine while the rest carries no information, and no global
denoising recovers what an opaque deposit removed. Occlusion is defined as a black rectangular mask
matching the partial-observation perturbation of closed-loop deployment
benchmarks~\cite{bench2driverobust}, so the two ladders are directly comparable at the overlap.

Every family is defined by a single physical parameter, and Appendix~\ref{supp:excluded} lists its value at each
of S1--S5 for all sixteen evaluated families, emitted directly from the released injector so that
specification and code cannot diverge. The ladder is therefore a specification rather than an ordinal label: a threshold at $s^*_{10}=2.5$ for
gaussian blur reads back as a defocus $\sigma$ of roughly $1.5$\,px. Endpoints were fixed before running
any model by one rule: S1 is a condition a deployed camera meets routinely, S5 is where a human would
call the view unusable, bracketing the knee rather than centering it. Snow and camera blackout have no natural scalar: their
parameter is an ordinal, and blackout ramps over camera identity rather than magnitude. Six further families the injector defines are excluded for measured or
structural reasons (Appendix~\ref{supp:excluded}).

\begin{figure*}[t]\centering
  \includegraphics[width=\textwidth]{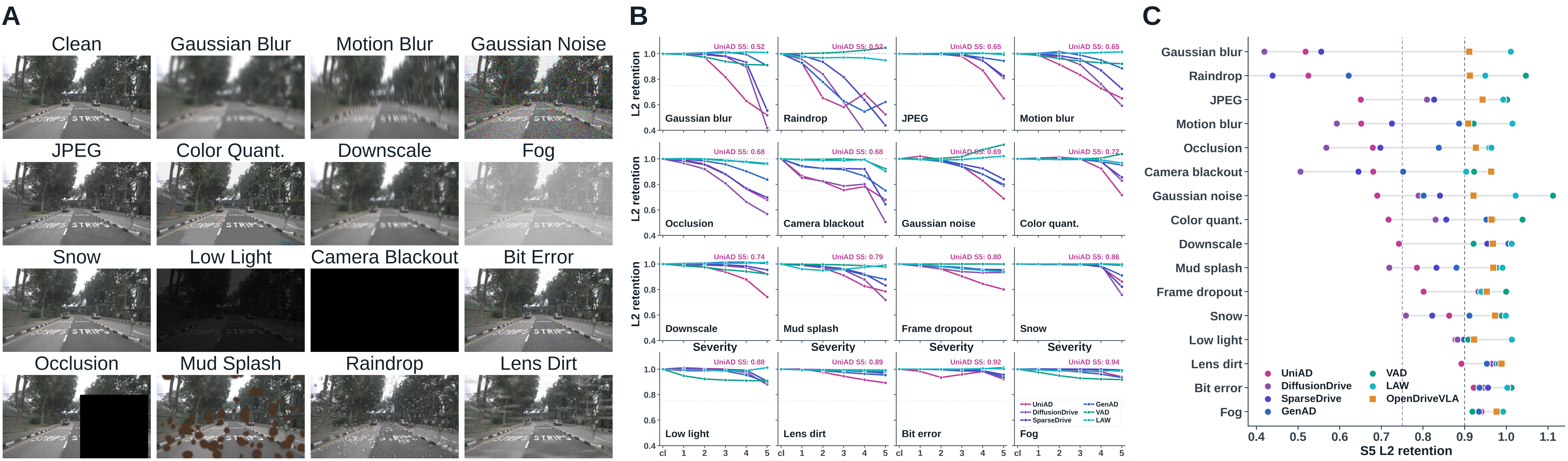}
  \caption{\textbf{Per-corruption image-degradation tolerance.} \textbf{(A)} The sixteen families on one nuScenes
  front-camera frame at S4, clean for reference; the bottom row is the four driving-specific families
  (Sec.~\ref{sec:driving}). Frame dropout has no single-frame rendering. \textbf{(B)} L2-retention
  curves for all six planners, ordered by UniAD fragility. \textbf{(C)} S5 retention for every
  corruption and planner. Blur, JPEG, and noise damage planning most; weather and bit-error are
  tolerated. OpenDriveVLA (squares) sits in $[0.909,0.989]$ on all sixteen.}
  \label{fig:tolerance}
\end{figure*}

\subsection{Models and protocols}
\label{sec:models}
Fifteen policies run under three protocols, each scored by its own released evaluator.

\noindent\textbf{nuScenes, open loop.} Six camera-input E2E planners span six architecture types:
UniAD (full-stack)~\cite{uniad}, VAD (vectorized)~\cite{vad}, SparseDrive (sparse)~\cite{sparsedrive},
GenAD (generative)~\cite{genad}, DiffusionDrive (diffusion)~\cite{diffusiondrive}, and LAW (latent
world model)~\cite{law}. Two newer designs run on the same data: Epona~\cite{epona}, an autoregressive
diffusion world model, and OpenDriveVLA-0.5B~\cite{opendrivevla}, a VLA planner. An ego-status-only
regressor in the style of AD-MLP~\cite{admlp} is an image-blind floor. LAW is trained without
perception \emph{supervision}, not blind: it consumes the surround images and responds when they are
removed (Sec.~\ref{sec:cross}). The six planners run inference on the full nuScenes validation set (6019 key-frames, six cameras);
trajectory L2 is averaged per each model's official protocol, over the 5119-frame valid-planning subset
for GenAD and DiffusionDrive and the full 6019 for the others. Those protocols differ, but the difference is inert for
retention: recomputing UniAD on the 5119-frame subset reproduces every retention we report to within
$0.0024$, two orders of magnitude below the effects we measure (Appendix~\ref{supp:subset}). Absolute L2 remains not strictly comparable across the two
groups (Sec.~\ref{sec:absolute}).

\noindent\textbf{NAVSIM, rule-based scoring.} Five planners run on NAVSIM~\cite{navsim,navsimv2}:
DrivoR~\cite{drivor}, camera-only on a DINOv2~\cite{dinov2} ViT-S backbone, GTRS-Aug~\cite{gtrs},
ZTRS~\cite{ztrs}, RAP~\cite{rap}, and iPad~\cite{ipad}. TransFuser~\cite{transfuser}, which fuses camera
with LiDAR, is a reference rather than a counted policy (Sec.~\ref{sec:pdms}).

\noindent\textbf{CARLA, closed loop.} SimLingo~\cite{simlingo} and TCP~\cite{tcp} are camera-only
agents, so a degraded camera cannot be covered by a second sensor (Sec.~\ref{sec:carla}).

\subsection{Metrics and tolerance threshold}
\label{sec:metrics}
DriveDegrade evaluates curves rather than a single score. We use retention rather than raw degraded L2
because the planners have different clean baselines and, in two cases, different official
valid-planning subsets; normalizing each degraded run by its paired clean run makes the primary
comparison the \emph{shape} of degradation. Absolute metrics are still reported where interpretable,
especially in the CARLA anchor.
Per (model, corruption, severity) we report trajectory L2 at 1/2/3\,s, vehicle and vulnerable-road-user
(VRU) collision, comfort, and an energy proxy. We define relative retention $r(s)=M(\text{clean})/M(s)$, where $M(s)$ is the model's mean L2 at
severity $s$ and $M(\text{clean})$ its clean-run value; for the reward-valued PDMS and EPDMS
(Sec.~\ref{sec:pdms},~\ref{sec:navsim4}) it is the reciprocal $M(s)/M(\text{clean})$, so higher is
always better and $1.000$ always means no measurable damage. We define the tolerance thresholds
$s^*_{10},s^*_{25}$ as the severity at which $r$ first drops below $0.90,0.75$. Open-loop evaluation is deterministic over a fixed per-model validation split, so each per-cell metric
is a population mean rather than a sample estimate: a frame-level bootstrap puts 95\% confidence
intervals below $0.01$ in retention. We therefore omit per-point error bars open-loop and report them
only for the stochastic CARLA anchor.
Appendix~\ref{supp:protocol} gives the protocol for adding a planner.

\section{Results}
\label{sec:results}

\subsection{Tolerance thresholds}
On UniAD (full nuScenes val, clean L2 $=0.94$\,m), mild degradation barely matters and each graded
image-quality corruption shows a clear knee (Fig.~\ref{fig:tolerance}B, Table~\ref{tab:thresh}). Among these
image-quality families, blur and noise are most fragile ($s^*_{10}\!\approx\!2.2$--$2.5$ for
motion/gaussian blur); weather and bit-error are tolerated across the whole range. Camera blackout is a
separate sensor-loss class rather than a graded degradation: it removes whole cameras, so its
$s^*_{10}=1.0$ (the first severity that drops a camera) is not comparable to the gradual thresholds and
we report it apart. Figure~\ref{fig:tolerance}C ranks the families by S5 retention across all six models.

\begin{table}[t]\centering\small
\caption{\textbf{Tolerance thresholds $s^*_{10}$} on full nuScenes val: the severity at which L2
retention first falls below $0.90$, linearly interpolated (higher is more tolerant). ``$>$5'' means
retention stays above $0.90$ through S5, and bold marks the planner that breaks first on each row. VAD
and LAW cross no threshold on any family, and fog and bit error break no planner, so those two columns
and two rows are omitted. Thresholds mostly rise from left to right, in order of decreasing camera
reliance, the ordering that Sec.~\ref{sec:cross} explains.}
\label{tab:thresh}
\setlength{\tabcolsep}{3pt}
\begin{tabular*}{\linewidth}{@{\extracolsep{\fill}}lcccc@{}}
\toprule
corruption & UniAD & DiffDr & SpDrv & GenAD \\
\midrule
\multicolumn{5}{@{}l}{\emph{graded image degradation}} \\
motion blur & \textbf{2.19} & 3.10 & 3.66 & 4.79 \\
gaussian blur & \textbf{2.47} & 3.96 & 4.09 & $>$5 \\
gaussian noise & \textbf{3.38} & 3.69 & 4.29 & 3.65 \\
jpeg & \textbf{3.72} & 4.35 & 4.38 & $>$5 \\
downscale & \textbf{3.64} & $>$5 & $>$5 & $>$5 \\
color quant. & \textbf{4.12} & 4.55 & 4.63 & $>$5 \\
frame dropout & \textbf{3.07} & $>$5 & $>$5 & $>$5 \\
snow & 4.67 & \textbf{4.41} & 4.52 & $>$5 \\
low light & \textbf{4.77} & 4.81 & $>$5 & 4.97 \\
\midrule
\multicolumn{5}{@{}l}{\emph{driving-specific contamination}} \\
raindrop & \textbf{1.10} & 1.49 & 2.31 & 1.19 \\
occlusion & 2.78 & \textbf{2.18} & 2.73 & 4.05 \\
mud splash & \textbf{3.13} & 3.71 & 4.25 & 4.42 \\
lens dirt & \textbf{4.68} & $>$5 & $>$5 & $>$5 \\
\midrule
\multicolumn{5}{@{}l}{\emph{sensor-loss stress test}} \\
camera blackout & \textbf{1.00} & \textbf{1.00} & 4.07 & 3.31 \\
\bottomrule
\end{tabular*}
\vspace{2pt}
\parbox{\linewidth}{\footnotesize\raggedright Camera blackout removes whole cameras rather than grading
an image parameter, so its threshold is not comparable to the graded families; $1.00$ means the mildest
setting already crosses. Planners: UniAD~\cite{uniad}, DiffDr = DiffusionDrive~\cite{diffusiondrive},
SpDrv = SparseDrive~\cite{sparsedrive}, GenAD~\cite{genad}, VAD~\cite{vad}, LAW~\cite{law}. The stricter
$s^*_{25}$ thresholds for all six planners are in Appendix~\ref{supp:excluded}, Table~\ref{tab:s25}.}
\end{table}

\subsection{Cross-model perception reliance}
\label{sec:cross}

\begin{figure}[b]\centering
  \includegraphics[width=\linewidth]{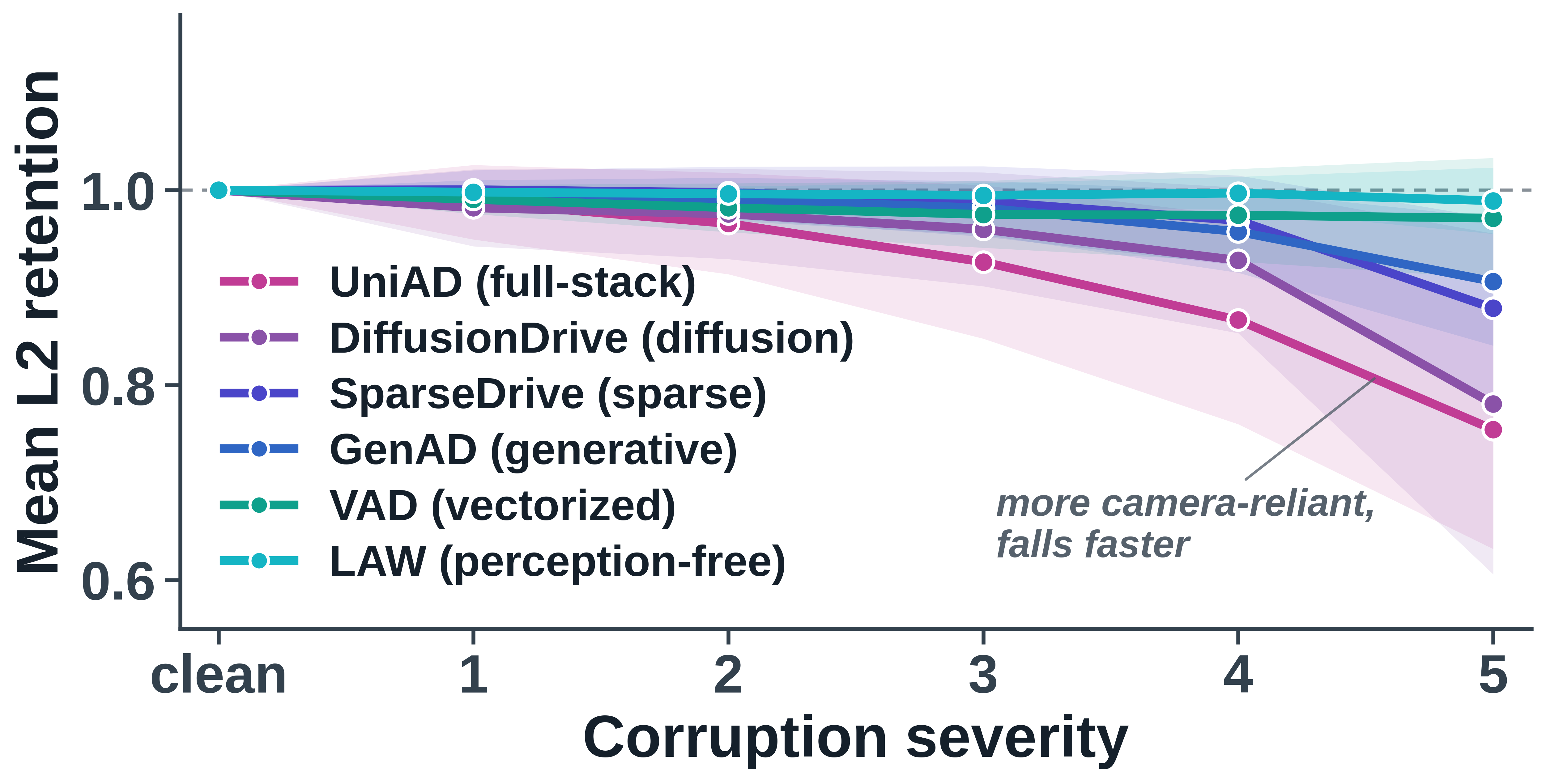}
  \caption{Mean L2 retention vs.\ severity. The slope indexes camera reliance: the full-stack and diffusion
  planners fall fastest, the vectorized and latent-world-model planners stay flat, an insensitivity to
  image quality that Table~\ref{tab:visrel} separates from disregard of the camera. The six solid curves
  average the twelve core corruptions; OpenDriveVLA (dashed) has full ladders for four of them, and the
  ego-only MLP (dotted) is $1.000$ everywhere by construction. On those four families the six planners
  run $0.625$ to $0.980$ at S5, so OpenDriveVLA's $0.932$ sits between VAD and LAW.}
  \label{fig:cross}
\end{figure}

Perception reliance differs sharply across architectures (Fig.~\ref{fig:cross},
Table~\ref{tab:visrel}). Averaged over the twelve corruptions at S5, mean L2 retention places the
full-stack and diffusion planners at the camera-reliant end, the sparse and generative ones in the
middle, and the vectorized and latent-world-model ones barely moving. The ordering holds per corruption
and is steepest under blur, where DiffusionDrive and UniAD keep only $0.42$ and $0.52$ of clean L2
(Fig.~\ref{fig:tolerance}B).

\noindent\textbf{Quality versus removal.} That spectrum is ambiguous at its flat end,
which is exactly where one would otherwise declare a winner. We split the corruption set by what it does
to the information in the image: quality families leave the scene present but worse, removal families
delete it. That supports an asymmetric test needing no extra runs. A planner that reads its camera must
lose accuracy when information is removed, whatever it does under quality loss, and one flat on both
axes is not using the camera.

Table~\ref{tab:visrel} applies it. None of the six is flat on both axes, so none ignores the camera
outright, though Sec.~\ref{sec:newmodels} adds a released VLA that very nearly is. UniAD,
DiffusionDrive, SparseDrive, and GenAD degrade on both. VAD and LAW are
nearly immune to quality ($0.97$ and $1.00$) yet still lose $7.7\%$ and $9.7\%$ when the front camera is
removed, and LAW declines monotonically under frame dropout ($0.996 \rightarrow 0.940$). Their flatness
is insensitivity to image \emph{quality}, not blindness: they use the coarse presence of visual evidence
but not its detail. That is the ego-status shortcut~\cite{bevplanner} stated precisely, and a weaker claim than a one-axis
reading would license. The gap between the axes is itself informative: largest for planners that
tolerate bad pixels but not missing ones, smallest where dependence on fine detail means degradation
destroys as effectively as deletion.

\begin{table}[t]\centering\footnotesize
\setlength{\tabcolsep}{2.5pt}
\caption{\textbf{Vision reliance on two axes.} S5 L2 retention under corruptions that \emph{degrade} the image
versus those that \emph{remove} it. A planner ignoring its camera would sit at $1.00$ in both columns;
none of the six does. VAD and LAW are quality-immune yet still lose accuracy when a camera is removed,
so their flat curves measure insensitivity to quality, not blindness. The VLA is the only entry with a
\emph{negative} gap. ``gap'' is quality minus blackout retention; bold marks the largest gap and the
largest loss under blackout.}
\label{tab:visrel}
\begin{tabular*}{\linewidth}{@{\extracolsep{\fill}}lccccc c@{}}
\toprule
& clean & \multicolumn{4}{c}{S5 retention\,$\uparrow$} & \\
\cmidrule(lr){3-6}
model & L2 (m)\,$\downarrow$ & mean & qual. & black. & drop. & gap \\
\midrule
UniAD~\cite{uniad}                   & 0.94 & 0.75 & 0.739 & 0.680 & 0.801 & 0.059 \\
DiffusionDrive~\cite{diffusiondrive} & 1.07 & 0.78 & 0.777 & \textbf{0.505} & 0.933 & \textbf{0.271} \\
SparseDrive~\cite{sparsedrive}       & 0.61 & 0.84 & 0.831 & 0.645 & 0.955 & 0.186 \\
GenAD~\cite{genad}                   & 0.91 & 0.91 & 0.915 & 0.752 & 0.948 & 0.163 \\
VAD~\cite{vad}                       & 0.97 & 0.97 & 0.969 & 0.923 & 0.999 & 0.047 \\
LAW~\cite{law}                       & 0.52 & 0.99 & 1.003 & 0.903 & 0.940 & 0.099 \\
\midrule
OpenDriveVLA~\cite{opendrivevla}     & 0.68 & 0.95 & 0.943 & 0.964 & 0.953 & $-0.021$ \\
ego-only MLP                         & 0.73 & 1.00 & 1.000 & 1.000 & 1.000 & 0.000 \\
\bottomrule
\end{tabular*}
\vspace{2pt}
\parbox{\linewidth}{\footnotesize\raggedright ``mean'' averages all twelve core S5 corruptions;
``quality'' the nine that degrade rather than remove (gaussian/motion blur, gaussian noise, JPEG, color
quantization, downscale, fog, snow, low light). Retention is clean L2 over degraded L2. Per-family
values for all sixteen families are in Fig.~\ref{fig:tolerance}C. The last row is our ego-status-only
regressor (no image input, Sec.~\ref{sec:scope}), $1.000$ everywhere by construction. OpenDriveVLA's
mean covers the eleven core families it reports at S5.}
\end{table}

\begin{figure}[t]\centering
  \includegraphics[width=\linewidth]{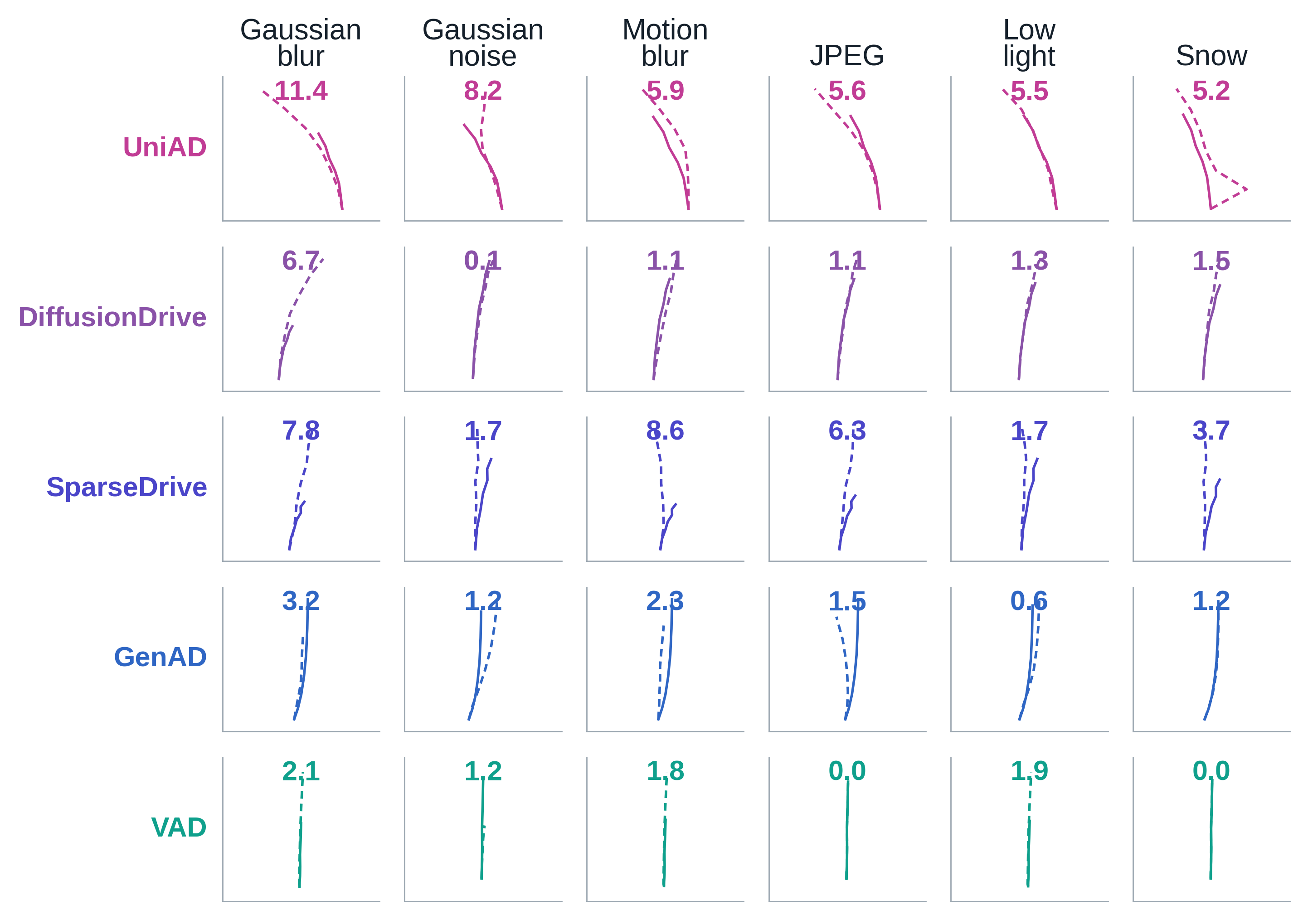}
  \caption{\textbf{Cross-method qualitative.} One nuScenes scene's planned ego trajectory for five planners (rows)
  under the six most disruptive families at S5 (columns, ordered by UniAD's endpoint shift): clean
  (solid) vs.\ corrupted (dashed), with the 3\,s endpoint shift in meters. Drift scales with camera
  reliance: UniAD and SparseDrive swing widely under blur, noise, and JPEG while VAD stays almost
  straight. LAW is omitted because its stored outputs are aggregate metrics, not trajectory points.}
  \label{fig:qual}
\end{figure}

\subsection{Absolute performance and rank reversal}
\label{sec:absolute}
Retention normalizes away clean accuracy, so Table~\ref{tab:visrel} also reports clean L2 and we release
every absolute cell. Two things follow that retention cannot express. First, degradation reorders the
leaderboard: GenAD beats VAD on clean data ($0.91$ against $0.97$\,m) and loses to it under gaussian
noise S5 ($1.13$ against $0.87$\,m), so a planner that is not state of the art clean can be the better
choice under degradation. Second, the planner trained without perception supervision has the lowest absolute L2 in \emph{every} condition tested: LAW is $0.52$\,m clean and
never exceeds $0.58$\,m anywhere. Our ego-status-only floor makes the same point from below, reaching
$0.73$\,m with no image at all, ahead of four of the five camera-supervised planners. On open-loop
nuScenes L2 no amount of visual supervision buys accuracy over a model leaning on ego
state~\cite{bevplanner,admlp}. Absolute L2 therefore cannot certify that a planner uses its camera well, which is why the degradation
\emph{slope} and the removal axis carry the diagnostic weight. Scaling every model against one reference, as RoboBEV~\cite{robobev} does for BEV detectors, assumes a
shared protocol. That holds for a single detection benchmark but not for planners shipping different
evaluators, horizons, and subsets, where the reference model's protocol becomes everyone's yardstick.
Releasing every absolute cell lets any baseline-relative score be recomputed.

\subsection{Path deviation, comfort, collision, and energy}
\label{sec:secondary}\label{sec:energy}
\begin{figure}[t]\centering
  \includegraphics[width=\linewidth]{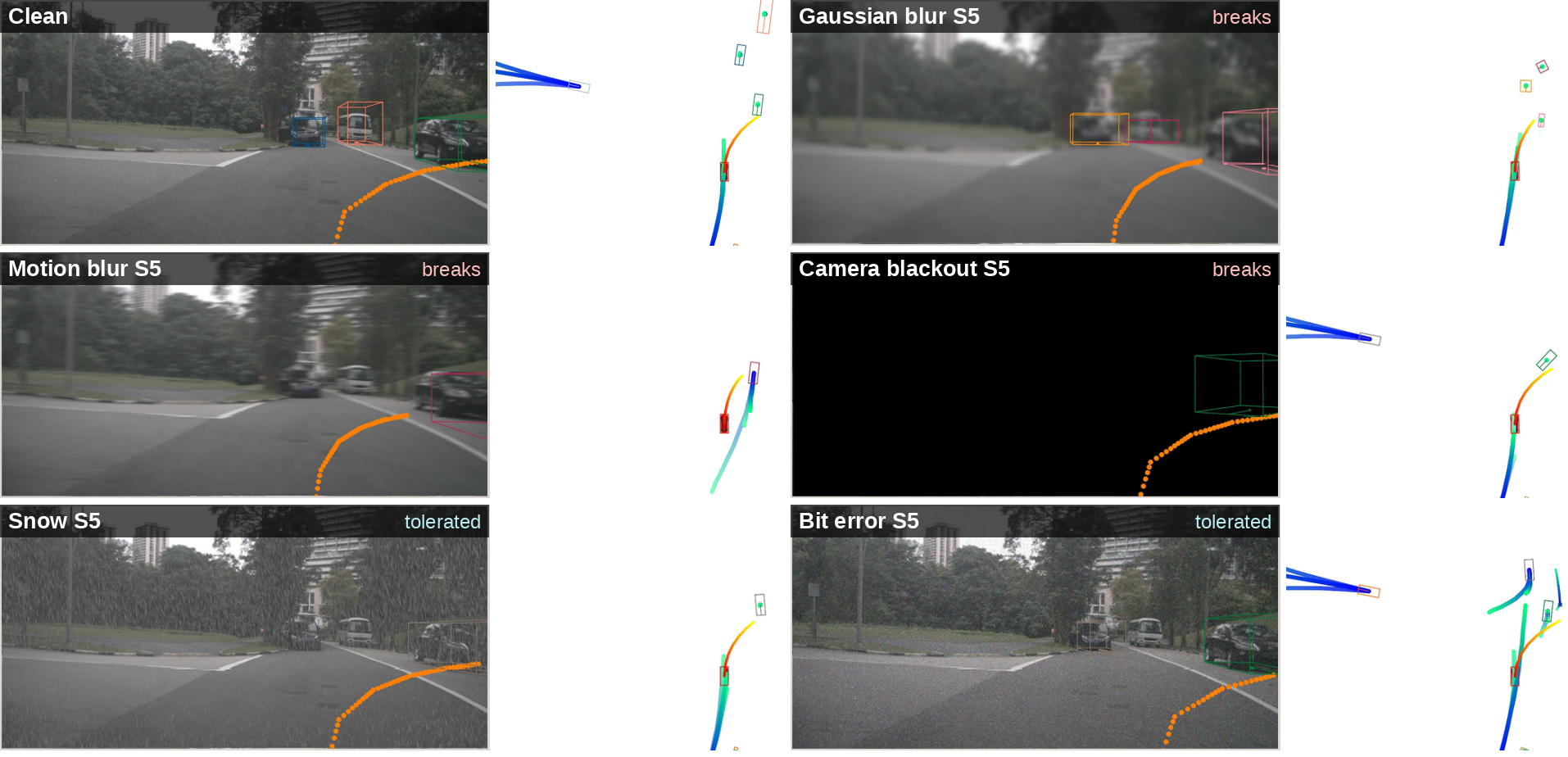}
  \caption{\textbf{Image evidence versus plan stability (UniAD, one scene).} Each cell pairs the front
  camera with UniAD's planned trajectory under one S5 corruption: clean, the three families that break
  the plan, and two that are tolerated. Blackout removes the image entirely yet the plan bends less than
  under blur, which is the asymmetry Sec.~\ref{sec:cross} measures. The full twelve-corruption montage
  with all six surround views is in Appendix~\ref{supp:montage}.}
  \label{fig:comparison}
\end{figure}

\begin{figure}[tb]\centering
  \includegraphics[width=\linewidth]{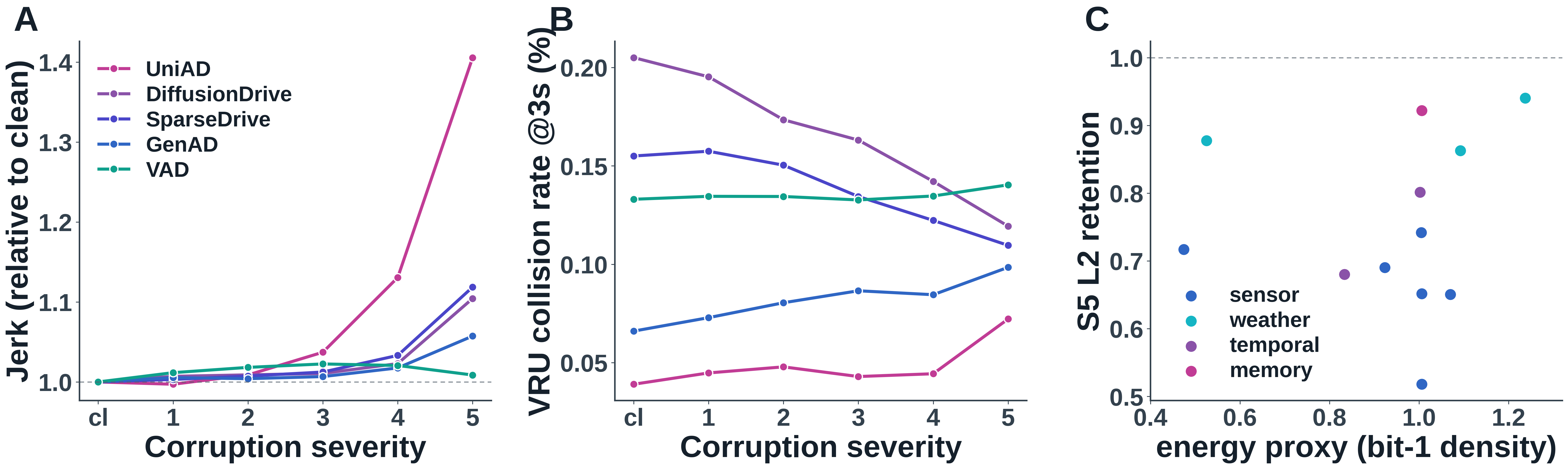}
  \caption{\textbf{Comfort, VRU collision, and energy across planners.} \textbf{(A)} Comfort: jerk relative to each planner's clean
  baseline, mean over the twelve corruptions. UniAD degrades most ($\sim$$1.4\times$ at S5) while VAD
  barely changes, mirroring the L2 spectrum. Absolute jerk is not comparable across planners, so (A) is
  normalized per planner. \textbf{(B)} VRU collision rate: small ($\sim$$0.02$--$0.25\%$) and roughly
  flat, a rare-event metric. \textbf{(C)} Energy proxy (bit-1 density) vs.\ S5 retention by corruption
  group (UniAD). ``cl'' = clean.}
  \label{fig:secondary}
\end{figure}

The planned path bends in proportion to camera reliance (Fig.~\ref{fig:qual}, Fig.~\ref{fig:comparison}): under gaussian blur S5 UniAD shifts its $3$\,s endpoint by $11.4$\,m and
DiffusionDrive by $6.7$\,m, while VAD barely moves ($2.1$\,m). Comfort and vehicle collision follow the
same ordering, UniAD's jerk rising from $1.92$ to $2.69$\,m/s$^3$ and its $3$\,s collision rate from
$0.65\%$ to $1.14\%$ (Fig.~\ref{fig:secondary}A). VRU collision does not: at $0.02$--$0.25\%$ it is a
rare-event metric, flat across severity for every model, because on a replayed scene the drifted plan
does not preferentially strike a VRU. The memory-interface energy proxy
(bit-1 density relative to clean, Fig.~\ref{fig:secondary}C) estimates the cost of moving and storing
image bytes~\cite{businvert,rahaddram,eden}, not system power: on an automotive system-on-chip the dominant term is
network computation, which degradation leaves unchanged. Within that scope, low light and color
quantization cut write energy while retaining $0.88$ and $0.72$.

\subsection{Driving-specific contamination}
\label{sec:driving}
The four contamination families (Fig.~\ref{fig:tolerance}A, bottom row) evaluate a failure mode the
generic taxonomy cannot represent: spatially concentrated damage. They reproduce the reliance spectrum of the generic taxonomy but reorder its fragile end
(Table~\ref{tab:thresh}, middle block). Raindrop is the most damaging family in the benchmark for the
camera-reliant planners, crossing $s^*_{10}$ between $1.10$ and $2.31$ where no generic family crosses
before $2.19$. GenAD is the sharpest case: it crosses only four thresholds among the twelve generic
families, yet raindrop breaks it barely past the mildest setting, so local refractive distortion defeats
a planner that global degradation hardly moves. Occlusion, the pure removal member, lands in the same
band as gaussian blur even though half of every frame stays pristine at S5, so these planners integrate
evidence across the full frame rather than from a region occlusion could spare. Epona reproduces the
raindrop result on its own protocol (Appendix~\ref{supp:newmodels}).

\subsection{Rule-based scoring on NAVSIM}
\label{sec:pdms}\label{sec:navsim4}
Open-loop L2 measures agreement with one recorded future. NAVSIM's PDMS instead rewards collision
avoidance, drivable-area compliance, and time-to-collision~\cite{navsim}, and EPDMS adds further driving
rules~\cite{navsimv2}. Six policies run over the same nine-cell slice (Table~\ref{tab:navsim}). They do
not share one protocol and we do not force one on them: GTRS-Aug and ZTRS publish on
\texttt{navhard\_two\_stage} under EPDMS, the other four on \texttt{navtest} ($12{,}146$ to $12{,}147$
scenes per cell) under PDMS, so the absolute row must not be pooled or ranked. Retention divides each model by its
own clean run, which is what makes the shapes comparable. Every clean cell reproduces its published
figure before any corrupted cell is read: ZTRS $45.45$ against $45.5$ and RAP $93.80$ against $93.8$
exactly, iPad within a tenth, GTRS-Aug $0.8$ EPDMS high.

The conclusions hold under rule-based scoring. TransFuser's PDMS falls under blur and further under
front-camera blackout, with every safety sub-score declining alongside it (Appendix~\ref{supp:pdms}). Its
retention is far milder than the camera-only planners' because LiDAR is untouched by image corruption,
which is why the main benchmark is camera-only: a sensor-rich agent understates the effect being
measured. Isolating the backbone from the modality, camera-only DrivoR on a DINOv2 ViT-S backbone matches
those retentions ($0.956$--$0.962$ at S5) with no second modality, so a foundation-model backbone alone
recovers what the ImageNet-pretrained camera-only planners lack. RAP, on a DINOv3~\cite{dinov3}
backbone, is almost untouched ($0.992$--$1.001$). The recovery stops at missing pixels: front-camera
blackout collapses DrivoR from $0.937$ to $0.531$ and costs RAP half its score and GTRS two thirds. A
foundation-model backbone therefore buys near-immunity to degraded pixels, not to missing ones.

\noindent\textbf{Null control.} Blackout S3 darkens the three rear cameras. ZTRS, GTRS-Aug, RAP, and
iPad each reproduce their clean score bit for bit, to fourteen decimal places: four confirmations, in
four separately patched devkits, that the injector is inside the loader and doing what the
specification says. TransFuser reproduces its clean PDMS and sub-scores in every reported digit; DrivoR
moves by $5\times10^{-4}$. Porting details are in Appendix~\ref{supp:navsim}.

\begin{table}[t]\centering\footnotesize
\caption{\textbf{NAVSIM under degradation}, six policies and nine cells each: ZTRS~\cite{ztrs},
GTRS-Aug~\cite{gtrs}, RAP~\cite{rap}, iPad~\cite{ipad}, DrivoR~\cite{drivor}, and TransFuser
(TF)~\cite{transfuser}. The clean row is the absolute score $\times100$; the rest are retention
(corrupted\,/\,clean, higher is better). ZTRS and GTRS score EPDMS on \texttt{navhard\_two\_stage}, the
other four PDMS on \texttt{navtest}, so absolute values are \emph{not} comparable across the rule. TF
fuses camera with LiDAR; DrivoR is camera-only on DINOv2 ViT-S. Blackout S3 darkens the three rear
cameras and is a null control. Bold: front-camera blackout costs every planner except TF between a
quarter and two thirds of its score. Sub-scores are in Appendix~\ref{supp:pdms}.}
\label{tab:navsim}\label{tab:pdms}\label{tab:navsim4}
\setlength{\tabcolsep}{2pt}
\begin{tabular*}{\linewidth}{@{\extracolsep{\fill}}lcc|cccc@{}}
\toprule
& \multicolumn{2}{c|}{EPDMS, \texttt{navhard}} & \multicolumn{4}{c}{PDMS, \texttt{navtest}} \\
\cmidrule(r){2-3}\cmidrule(l){4-7}
condition & ZTRS & GTRS & RAP & iPad & DrivoR & TF \\
\midrule
clean (abs., $\times$100) & 45.5 & 42.9 & 93.8 & 91.8 & 93.7 & 83.4 \\
\midrule
gauss.\ blur S3      & 0.958 & 1.002 & 1.000 & 0.990 & 0.998 & 0.994 \\
gauss.\ blur S5      & 0.882 & 0.939 & 0.998 & 0.947 & 0.961 & 0.956 \\
motion blur S3       & 0.943 & 0.963 & 1.001 & 0.990 & 0.986 & 0.991 \\
motion blur S5       & 0.927 & 0.903 & 0.998 & 0.961 & 0.962 & 0.965 \\
jpeg S3              & 0.963 & 0.961 & 0.999 & 0.995 & 0.995 & 1.003 \\
jpeg S5              & 0.787 & 0.862 & 0.992 & 0.963 & 0.956 & 0.968 \\
\midrule
blackout S3 (rear)   & 1.000 & 1.000 & 1.000 & 1.000 & 1.001 & 1.000 \\
blackout S5 (front)  & \textbf{0.462} & \textbf{0.351} & \textbf{0.494} & \textbf{0.773} & \textbf{0.567} & \textbf{0.903} \\
\bottomrule
\end{tabular*}
\end{table}

\subsection{A world model and a vision-language-action planner}
\label{sec:newmodels}
World models and VLA planners are two families proposed as successors to the imitation-trained
planners above. We ran one of each: \emph{Epona}~\cite{epona}, an autoregressive diffusion world model,
and \emph{OpenDriveVLA-0.5B}~\cite{opendrivevla}. Neither buys robustness: Epona shows the
camera-reliant profile and reproduces our raindrop finding independently. OpenDriveVLA is the
informative case: its clean L2 reproduces the published $0.35$\,m under its own ST-P3-style metric ($0.68$\,m under
the UniAD metric of Table~\ref{tab:visrel}), yet no cell in a forty-eight-cell
sweep moves it, every retention falling in $[0.909, 0.997]$, and the front camera going dark costs $3.6\%$, so it is flat on both
axes and alone in having a \emph{negative} gap (Table~\ref{tab:visrel}). Zeroing all six views on the full validation set still
leaves retention $0.885$: blinding it entirely costs $11.5\%$, where UniAD loses $32\%$ from the front
camera alone. Its open-loop accuracy is carried by ego status and the command: the AD-MLP and BEV-Planner
critique~\cite{admlp,bevplanner} inside a VLA, found by a corruption ladder rather than by ablating an
input. Details in Appendix~\ref{supp:newmodels}.

\subsection{Closed-loop anchor in CARLA}
\label{sec:carla}
\begin{figure}[t]\centering
  \includegraphics[width=\linewidth]{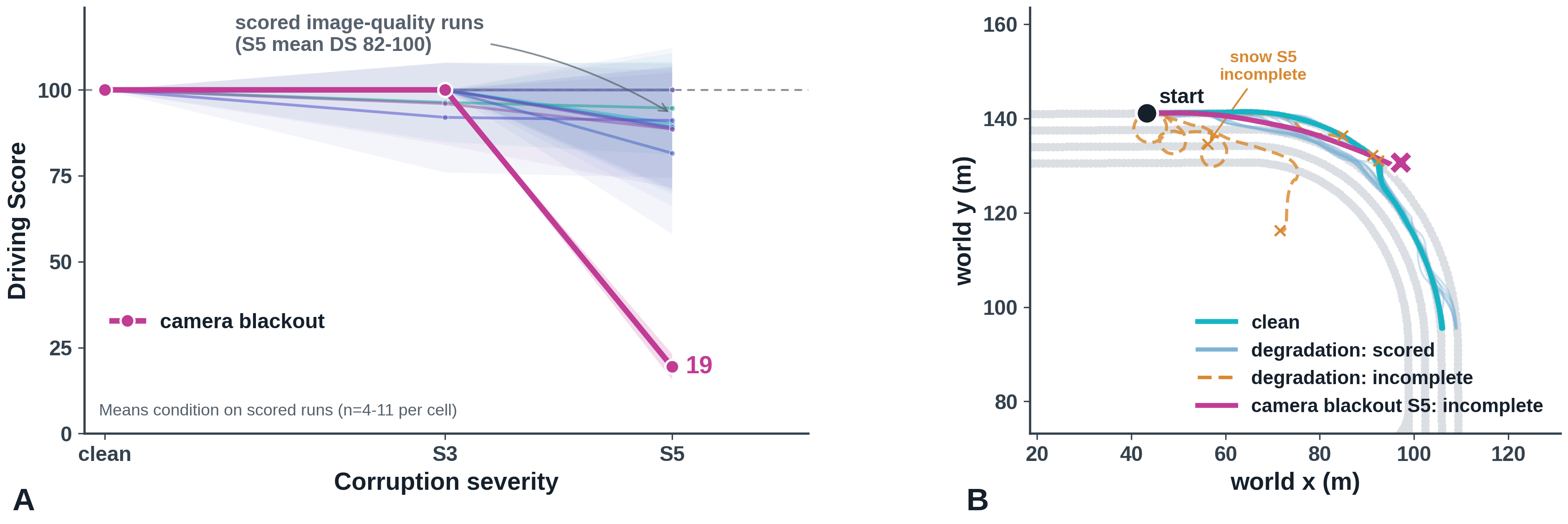}
  \caption{\textbf{CARLA closed-loop evidence and run status.} \textbf{(A)} Complete-case DS on Town10;
  bands are $\pm1$ std over scored runs. \textbf{(B)} Route-148 traces: solid blue is scored
  degradation, dashed orange incomplete, magenta incomplete blackout S5. The snow-S5 loop is excluded
  from the DS mean, which the identification bounds account for.}
  \label{fig:carla_anchor}
\end{figure}

Camera-only agents anchor the trends under feedback, since a fusion planner would cover a degraded
camera with LiDAR. With all twelve families injected into SimLingo's~\cite{simlingo} camera stream on
Town10, front-camera blackout at S5 reduces mean Driving Score (DS) from $100$ to $\mathbf{19.4}$ while
image-quality conditions stay above $81$ (Table~\ref{tab:carla}). Complete-case averaging is optimistic
exactly where the benchmark bites, so we bound the unscored runs rather than drop them: assigning them
any DS in $[0,100]$ puts blackout S5 in $[15.6,35.6]$ and the widest image-quality interval in
$[53.5,93.5]$, so the availability-versus-quality separation survives any assignment (Fig.~\ref{fig:carla_anchor}). Rankings
\emph{within} the image-quality group do not. Repeating the slice on Town03 and Town05 reproduces the
ordering in both, with collision counts an order of magnitude above any other cell.

\noindent\textbf{Second agent and null control.} One agent cannot carry a generalization, so we ran
the identical Town05 slice with TCP~\cite{tcp}. The slice contains a \emph{free null control}: blackout
S3 names the three rear views and neither agent has one, so it changes no pixel and measures how far a
driving score wanders on rerun with the input untouched. SimLingo's null does not move, all eight runs
reproducing clean exactly, so its blackout-S5 collapse is unambiguous. TCP's moves four of twelve runs
at mean $-12.8$ DS, as large as any effect we measure on it, so \emph{no TCP cell separates from its own
null} and we report that instead of per-cell means. Blackout is also not the same treatment for the two,
since SimLingo goes blind while TCP loses the center third of a panorama.

\noindent\textbf{Contamination families.} Run closed-loop on Town03, the open-loop ordering survives at
the ends and collapses in the middle: lens dirt stays mildest while occlusion, mud splash, and raindrop converge at
S5 to within $1.3$ DS of each other, where open loop separates them by more than a severity step. What differs is the
\emph{distribution} of the loss. Occlusion spreads it across five of eight runs; raindrop and mud splash
leave six runs untouched at $100$ and annihilate both seeds of the one route already below par. Deleting
part of the view costs a little almost everywhere, whereas refractive distortion costs nothing until the
route demands what it hides, and a per-cell mean reports the two as the same number. Six runs end on the
blocked-actor criterion, all in severe cells; counting only completed runs would report raindrop S5 at
$86.7$ rather than $65.6$. Route traces and per-run scores are in Appendix~\ref{supp:carla}.

\begin{table}[t]\centering\small
\caption{\textbf{CARLA closed-loop Driving Score (DS)}, higher is better. Top: the seven-cell slice on
three towns for SimLingo~\cite{simlingo} and, last column, TCP~\cite{tcp} on Town05. Bottom: SimLingo
on the remaining Town10 corruptions and on the contamination families on Town03, as S3\,/\,S5.
Blackout S3 is the null control of Sec.~\ref{sec:carla}: SimLingo's does not move, TCP's moves four of
twelve runs. DS conditions on completed scored runs; bold marks the front-camera blackout collapse.}
\label{tab:carla}
\setlength{\tabcolsep}{3pt}
\begin{tabular*}{\linewidth}{@{\extracolsep{\fill}}lccc|c@{}}
\toprule
& \multicolumn{3}{c|}{SimLingo} & TCP \\
\cmidrule(r){2-4}\cmidrule(l){5-5}
condition & Town10 & Town03 & Town05 & Town05 \\
\midrule
clean               & 100.0 & 92.5 & 90.0 & 68.8 \\
gaussian blur S3    & 100.0 & 92.5 & 95.0 & 68.7 \\
gaussian blur S5    & 90.2  & 72.0 & 85.0 & 55.1 \\
motion blur S3      & 96.4  & 96.2 & 95.0 & 71.7 \\
motion blur S5      & 94.7  & 92.5 & 87.0 & 68.2 \\
blackout S3 (null)  & 100.0 & 92.5 & 90.0 & 57.8 \\
blackout S5 (front) & \textbf{19.4} & \textbf{39.8} & \textbf{36.3} & 60.9 \\
\bottomrule
\end{tabular*}

\vspace{3pt}
\begin{tabular*}{\linewidth}{@{\extracolsep{\fill}}lc@{\hspace{8pt}}lc@{}}
\toprule
\multicolumn{4}{@{}l}{\emph{SimLingo, remaining corruptions, Town10, S3\,/\,S5}} \\
\midrule
fog           & 100.0\,/\,100.0 & gaussian noise & 100.0\,/\,89.1 \\
bit error     & 100.0\,/\,100.0 & low light      & 100.0\,/\,88.6 \\
frame dropout & 100.0\,/\,100.0 & color quant    & 96.0\,/\,88.6 \\
jpeg          & 92.0\,/\,91.1   & downscale      & 100.0\,/\,81.6 \\
snow          & 100.0\,/\,89.2  &                &                \\
\midrule
\multicolumn{4}{@{}l}{\emph{SimLingo, contamination families, Town03, S3\,/\,S5}} \\
\midrule
lens dirt     & 92.5\,/\,87.5   & raindrop       & 89.9\,/\,65.6 \\
mud splash    & 92.5\,/\,64.5   & occlusion      & 84.3\,/\,64.3 \\
\bottomrule
\end{tabular*}
\end{table}


\section{Evaluation Scope and Limitations}
\label{sec:scope}
DriveDegrade supports comparative claims under controlled conditions: model, split, evaluator, and
operator are fixed, and only the camera stream is degraded. It measures tolerance thresholds and relative vision
reliance, not deployment safety. Because open-loop nuScenes rewards ego-status
shortcuts, a flat curve is not by itself evidence of quality; the quality-versus-removal split of
Sec.~\ref{sec:cross} separates robustness from indifference. None of the six planners is fully non-visual, so we pin the blind floor with an ego-status-only
regressor in the style of AD-MLP~\cite{admlp}: a planner whose blackout-S5 accuracy stays above that
floor is extracting value from its remaining cameras.

Four limitations follow. Open-loop nuScenes measures replayed planning error, so collision and comfort
miss feedback compounding. The closed-loop anchor spans two agents and three towns but remains one simulator with few routes; the
identification bounds keep the blackout separation robust to unscored outcomes but do not justify
fine-grained ranking among image-quality conditions, and TCP's null moves as much as its largest
effect. The corruptions are controlled synthetic stressors, not all real sensor
physics, and the energy figure is an analytical proxy rather than measured watts.

\section{Conclusion}
DriveDegrade turns camera degradation from a perception-only robustness question into a planning-level
tolerance measurement. Across sixteen corruption families, with one injector reaching fifteen policies,
it locates clear breakpoints for blur, JPEG, noise, and camera loss while weather and bit-error are
often tolerated. The same curves reveal vision reliance that clean-set scores hide, and the CARLA anchor
confirms camera loss remains load-bearing under feedback.

\noindent\textbf{Artifact.} The injector, hooks, configs, scripts, and per-cell CSVs are released at
\url{https://github.com/UGA-MOBILITY-LAB/DriveDegrade}, MIT licensed and with no raw sensor
data. Every cell is traceable to a checkpoint, split, corruption, severity, and seed, and every table
and figure regenerates from them on CPU. We version the specification and treat operator changes as
major versions, so published numbers stay valid (Appendix~\ref{supp:protocol}).

\noindent\textbf{Acknowledgments.} This study was supported in part by resources and technical expertise
from the Georgia Advanced Computing Resource Center, a partnership between the University of Georgia's
Office of the Vice President for Research and Office of the Vice President for Information Technology.

{\small\bibliographystyle{ieeenat_fullname}\bibliography{main}}

\clearpage
\appendix
\section*{Appendix}
\noindent This appendix holds the measurements and derivations that the paper cites but has no room to
spell out. All numbers are reproducible from the released artifact, and the script that regenerates each
is named at its point of use.

\section{The two nuScenes evaluation subsets}
\label{supp:subset}
Sec.~\ref{sec:models} states that the 5119-frame and 6019-frame averaging protocols differ but
that the difference is inert for retention. This is the measurement behind that claim.

The subset is exactly reconstructible. The 5119 frames are the 6019 key-frames minus the last six of
each of the 150 validation scenes, which are the frames where the ego has no complete $3$\,s future.
That is 900 frames, and our reconstruction returns 5119 exactly. The script asserts the count before
writing anything (\texttt{analysis/valid\_planning\_tokens.py}).

Recomputing UniAD's per-frame L2 from its stored predictions and restricting to those tokens reproduces
every retention we report to within $0.0024$: gaussian blur S5 gives $0.519$ recomputed against $0.518$
reported, motion blur S5 $0.652$ against $0.652$, JPEG S5 $0.653$ against $0.650$, camera blackout S5
$0.678$ against $0.680$, and raindrop S5 $0.524$ against $0.525$. Averaging the same predictions over
all 6019 frames instead inflates those retentions by up to $0.24$, since gaussian blur S5 would read
$0.749$. The 900 excluded frames score against a padded target and contribute meaningless error.

We also checked the mechanism in the evaluator source rather than inferring it. UniAD calls
\texttt{planning\_metrics(pred[:, :6, :2], gt[:, :6, :2], sdc\_planning\_mask[:, :6, :2], seg)}, which
is a per-timestep validity mask rather than a frame filter. A frame three from the end of a scene
therefore still contributes the horizons it does have, while GenAD and DiffusionDrive discard it
entirely. The two protocols are not equivalent and we do not claim they are. What the measurement
establishes is narrower and sufficient: moving UniAD onto the other group's frame set changes its
retention by at most $0.0024$, two orders of magnitude below the $0.1$ to $0.5$ per-family effects we
report. Absolute L2 remains not strictly comparable across the groups, which is why the primary analysis
is retention. Per-cell numbers are in \texttt{results/analysis/subset\_check.csv} and the recomputation
is \texttt{analysis/subset\_check.py}.

\section{Corruption specification and excluded families}
\label{supp:excluded}
Table~\ref{tab:spec} gives the physical parameter of every evaluated family at each severity.

\begin{table}[t]\centering\footnotesize
\caption{\textbf{Corruption specification.} Each family is defined by one physical parameter whose
value the severity ladder S1--S5 lists directly, so an ordinal threshold always reads back as a
physical quantity. Endpoints follow one rule, fixed before any model was run: S1 is a condition a
deployed camera meets routinely, S5 is where a human would call the view unusable, bracketing the
knee rather than centering the ladder on it. Snow and camera blackout have no natural scalar: snow is
an ordinal over an internal parameter set, and blackout ramps over camera identity (S5 = front
camera). The first twelve families are the core set; the last four are the driving-specific additions. Emitted from the released injector by \texttt{analysis/emit\_spec\_table.py}.}
\label{tab:spec}
\setlength{\tabcolsep}{2.5pt}
\begin{tabular*}{\linewidth}{@{\extracolsep{\fill}}llccccc@{}}
\toprule
family & parameter & S1 & S2 & S3 & S4 & S5 \\
\midrule
gaussian blur & defocus $\sigma$ (px) & 0.5 & 1 & 2 & 4 & 7 \\
motion blur & kernel length (px) & 3 & 7 & 11 & 17 & 25 \\
gaussian noise & noise $\sigma$ (8-bit) & 5 & 12 & 25 & 45 & 80 \\
JPEG & encoder quality & 50 & 30 & 20 & 12 & 6 \\
color quant. & bits/channel & 6 & 5 & 4 & 3 & 2 \\
downscale & resize factor & 0.75 & 0.5 & 0.33 & 0.25 & 0.15 \\
fog & transmission $t$ & 0.85 & 0.7 & 0.55 & 0.45 & 0.35 \\
snow & ordinal & 1 & 2 & 3 & 4 & 5 \\
low light & brightness gain & 0.7 & 0.5 & 0.35 & 0.2 & 0.1 \\
frame dropout & $P(\text{stale})$ & 0.1 & 0.25 & 0.5 & 0.75 & 0.9 \\
cam.\ blackout & cameras removed & 1 & 2 & 3 & 4 & front \\
bit error & flip probability & 0.0001 & 0.0005 & 0.002 & 0.008 & 0.03 \\
\midrule
\multicolumn{7}{@{}l}{\emph{driving-specific families (new)}} \\
occlusion & area fraction & 0.05 & 0.1 & 0.2 & 0.35 & 0.5 \\
mud splash & covered fraction & 0.02 & 0.05 & 0.1 & 0.2 & 0.35 \\
raindrop & droplet count & 30 & 80 & 180 & 400 & 800 \\
lens dirt & film opacity & 0.2 & 0.35 & 0.55 & 0.75 & 0.92 \\
\bottomrule
\end{tabular*}
\end{table}

\noindent\textbf{Stricter thresholds.} Table~\ref{tab:thresh} reports $s^*_{10}$.
Table~\ref{tab:s25} gives the stricter $s^*_{25}$, the severity at which retention first falls below
$0.75$, for all six planners. Regenerate with \texttt{analysis/absolute\_and\_thresholds.py}.

\begin{table}[t]\centering\small
\caption{\textbf{Stricter tolerance thresholds $s^*_{25}$} on full nuScenes val, linearly interpolated
(higher is more tolerant). ``$>$5'' means retention stays above $0.75$ through S5, and bold marks the
planner that breaks first on each row. VAD and LAW cross no $s^*_{25}$ threshold, and frame dropout,
snow, low light, fog, bit error, and lens dirt break no planner at this level, so those columns and
rows are omitted. DiffDr = DiffusionDrive~\cite{diffusiondrive}, SpDrv = SparseDrive~\cite{sparsedrive}.}
\label{tab:s25}
\setlength{\tabcolsep}{3pt}
\begin{tabular*}{\linewidth}{@{\extracolsep{\fill}}lcccc@{}}
\toprule
corruption & UniAD~\cite{uniad} & DiffDr & SpDrv & GenAD~\cite{genad} \\
\midrule
\multicolumn{5}{@{}l}{\emph{graded image degradation}} \\
motion blur & \textbf{3.80} & 4.08 & 4.83 & $>$5 \\
gaussian blur & \textbf{3.36} & 4.31 & 4.48 & $>$5 \\
gaussian noise & \textbf{4.57} & $>$5 & $>$5 & $>$5 \\
jpeg & \textbf{4.54} & $>$5 & $>$5 & $>$5 \\
downscale & \textbf{4.94} & $>$5 & $>$5 & $>$5 \\
color quant. & \textbf{4.84} & $>$5 & $>$5 & $>$5 \\
\midrule
\multicolumn{5}{@{}l}{\emph{driving-specific contamination}} \\
raindrop & \textbf{1.65} & 2.41 & 3.38 & 2.19 \\
occlusion & 4.16 & \textbf{3.40} & 4.27 & $>$5 \\
mud splash & $>$5 & \textbf{4.80} & $>$5 & $>$5 \\
\midrule
\multicolumn{5}{@{}l}{\emph{sensor-loss stress test}} \\
camera blackout & 4.32 & \textbf{4.17} & 4.62 & $>$5 \\
\bottomrule
\end{tabular*}
\end{table}

\noindent\textbf{Excluded families.} The injector defines six families that the study does not evaluate. Two are excluded on measurement and
four on structure.

We ran both measurable candidates through the full UniAD ladder rather than asserting they were
uninteresting. Contrast is flat, with L2 between $0.94$ and $1.01$\,m across S1 to S5 against a
$0.94$\,m clean baseline, so it admits no threshold to report. Impulse noise is non-monotonic in
severity under this injector's parameterization, at $1.41$\,m for S1, $1.30$\,m for S3, and $1.64$\,m
for S5, which violates the single-knob severity contract the ladder rests on. Gaussian noise is the
well-behaved representative of the same sensor-noise axis and is evaluated in its place.

The four structural exclusions are these. Frost and spatter model on-lens ice and water deposits, and
both are superseded by the vehicle-specific raindrop and lens-dirt families of Sec.~\ref{sec:driving}, which are
parameterized by a physical occluding quantity rather than an ordinal. Zoom blur is the radial special
case of the motion-blur axis already evaluated. Latency is a timing fault rather than an image fault:
under open-loop replay it reduces to a frame offset, which frame dropout already covers, and its
deployment effect is exercised by the closed-loop anchor.

\section{Injector seeding and realization variance}
\label{supp:seeding}
Six families are deterministic functions of the frame and reach every model bit-identically: gaussian
blur, JPEG, color quantization, downscale, fog, and camera blackout. Motion blur, gaussian noise, snow,
low light, bit error, and frame dropout draw a per-frame random realization. The released injector
derives each seed as $\mathrm{blake2b}(\text{seed}, \text{frame\_id}, \text{camera})$, so every
realization is reproducible and identical across models.

The stochastic cells reported in the paper were drawn with a process-local string hash, so their exact
realizations cannot be regenerated, and one model's copy of the injector used an older, statistically
equivalent \texttt{bit\_error} implementation. What every model shared is the operator and its severity
parameter, and each cell averages over 6019 frames and six cameras. We measured the realization variance
directly. Re-running the two most
stochastic cells across three seeds shifts retention by $0.004$ for gaussian noise S5 and $0.002$ for
snow S5, against per-family effects of $0.1$ to $0.5$. The seed choice is inert for every conclusion in
the paper (\texttt{analysis/seed\_check.py}, \texttt{results/seed\_check.csv}).

\section{The PDMS slice}
\label{supp:pdms}
Table~\ref{tab:navsim} reports PDMS retention for TransFuser~\cite{transfuser}
and DrivoR~\cite{drivor} on NAVSIM \texttt{navtest}; Table~\ref{tab:pdms_sub} gives the sub-scores. Two
details belong with them.

First, blacking out the rear cameras at S3 leaves TransFuser's PDMS and every sub-score unchanged in
every reported digit, at $0.834$. DrivoR moves from $0.93691$ to $0.93738$, a change of
$5\times10^{-4}$ that is two orders of magnitude below any effect we report. The null control is
therefore exact for TransFuser and for the four planners of Sec.~\ref{sec:navsim4},
and close to exact for DrivoR.

\begin{table}[t]\centering\small
\caption{\textbf{PDMS sub-scores under degradation} (NAVSIM \texttt{navtest}, $12{,}146$ to $12{,}147$
scenes per cell), for TransFuser~\cite{transfuser}, which fuses camera with LiDAR, and
DrivoR~\cite{drivor}, camera-only on a DINOv2 ViT-S backbone. NC = no collision, DAC = drivable-area
compliance, TTC = time-to-collision, EP = ego progress; all higher is better. Degradation lowers PDMS
and every safety sub-score, so the effect is not an artifact of imitation L2.}
\label{tab:pdms_sub}
\setlength{\tabcolsep}{3pt}
\begin{tabular*}{\linewidth}{@{\extracolsep{\fill}}lcccccc@{}}
\toprule
condition & PDMS & ret. & NC & DAC & TTC & EP \\
\midrule
\multicolumn{7}{@{}l}{\emph{TransFuser (camera+LiDAR)}}\\
clean            & 0.834 & 1.000 & 0.978 & 0.921 & 0.928 & 0.786 \\
jpeg S5          & 0.807 & 0.968 & 0.968 & 0.906 & 0.903 & 0.766 \\
motion blur S5   & 0.805 & 0.965 & 0.968 & 0.911 & 0.905 & 0.755 \\
gaussian blur S5 & 0.797 & 0.956 & 0.966 & 0.905 & 0.905 & 0.741 \\
blackout S3 (rear) & 0.834 & 1.000 & 0.978 & 0.921 & 0.928 & 0.786 \\
blackout S5 (front) & 0.753 & 0.903 & 0.933 & 0.886 & 0.845 & 0.724 \\
\midrule
\multicolumn{7}{@{}l}{\emph{DrivoR (camera-only, DINOv2 ViT-S)}}\\
clean            & 0.937 & 1.000 & 0.990 & 0.989 & 0.967 & 0.899 \\
jpeg S5          & 0.896 & 0.956 & 0.985 & 0.962 & 0.953 & 0.853 \\
motion blur S5   & 0.901 & 0.962 & 0.980 & 0.977 & 0.941 & 0.860 \\
gaussian blur S5 & 0.900 & 0.961 & 0.979 & 0.972 & 0.925 & 0.880 \\
blackout S3 (rear) & 0.937 & 1.001 & 0.990 & 0.989 & 0.967 & 0.900 \\
blackout S5 (front) & 0.531 & 0.567 & 0.832 & 0.697 & 0.757 & 0.479 \\
\bottomrule
\end{tabular*}
\end{table}

Second, every safety sub-score declines with PDMS rather than the aggregate alone: no-collision,
drivable-area compliance, and time-to-collision all fall under blur S5 and further under front-camera
blackout. The damage is therefore not confined to imitation distance, which is the point of running the
slice at all.

The slice covers two agents and four corruptions at two severities. It is a control, not a second
benchmark. Raw per-cell values are in \texttt{results/navsim/pdms\_summary.csv}.

\section{Porting the injector into the NAVSIM planners}
\label{supp:navsim}
Sec.~\ref{sec:navsim4} reports thirty-six cells across four NAVSIM planners,
ZTRS~\cite{ztrs}, GTRS-Aug~\cite{gtrs}, RAP~\cite{rap}, and iPad~\cite{ipad}, whose checkpoints needed
substantial adaptation before they would run inference. This section records what that involved, because
it bears on how much the resulting numbers should be trusted.

Every NAVSIM-era planner ships its own fork of the devkit, and each loads camera images at the same
place, \texttt{Camera.from\_camera\_dict}. Corrupting there means the model is scored under exactly the
same operator as every other model in the benchmark, with no corrupted copies on disk. Copying a patched
file between forks is unsafe once upstream diverges, so the released porter inserts the block instead
and refuses to patch when the call site is not found verbatim
(\texttt{integration/navsim\_inject\_patch.py}). A no-op injector reads as perfect robustness, which is
the one failure mode this benchmark cannot afford. RAP is the interesting case: it loads a second,
rendered image that is a 3D rasterization rather than a camera capture, and only the real camera stream
is corrupted, since degrading a synthetic view would not correspond to any sensor fault.

\noindent\textbf{GTRS-Aug.} Its released \texttt{predict\_step} dispatches only on \texttt{GTRSAgent}
and \texttt{DPAgent}, while \texttt{GTRSAugAgent} subclasses \texttt{AbstractAgent} in both public
checkouts, so the agent does not run out of the box. Three patches restore the routing and the feature-computation
signature, after which the clean run reproduces the published score. Reproducing the published number is
what distinguishes restoring missing wiring from inventing upstream behavior, and it is why we report
the agent rather than excluding it.

\noindent\textbf{RAP.} Seven fixes were needed and all share one root cause: the released checkpoint is
saved in training configuration and the inference path was never exercised upstream. The scorer and the
training metric cache have to be disabled or redirected, the trajectory sampling horizon has to be set
to the value the checkpoint was trained with rather than the config default, the domain-adversarial
distillation head has to be turned off, and the state-dict load has to tolerate exactly the missing
classifier keys and no others. The model's forward pass also reads two attributes that exist only during
training, so the whole adversarial block is guarded rather than patched attribute by attribute.

\noindent\textbf{Validation gate.} Because so much upstream code was adapted, no corrupted cell is read
until the clean cell reproduces the published figure. ZTRS gives $0.4545$ against a published $45.5$ and
RAP gives $0.9380$ against $93.8$, both exact. iPad gives $0.9179$ against a reported $91.7$. GTRS-Aug
gives $0.4294$ against a published $42.1$, which is $0.8$ EPDMS high, and we report the discrepancy
rather than round it away. All ports and fixes are released.

\noindent\textbf{The null control, at full precision.} Blackout S3 names the three rear cameras. Each of
the four planners reproduces its clean score to fourteen decimal places, for example RAP at
$0.9380278397075544$ in both the clean and the blackout-S3 cell. The aggregation script flags any
retention of exactly $1.0$ for this reason
(\texttt{results/navsim/navsim\_degrade\_aggregate.py}). That script also hard-codes the retention
direction, since PDMS and EPDMS are rewards and L2 is an error: applying the L2 orientation to a score
would report every degradation as an improvement without raising an error.

\section{Epona and OpenDriveVLA}
\label{supp:newmodels}
\noindent\textbf{Epona.} Table~\ref{tab:epona} gives the Epona~\cite{epona} cells summarized in the
Sec.~\ref{sec:newmodels}. Two caveats belong with them. Epona conditions on a single
front camera, so camera-blackout severities below S5, which darken only rear views, are clean by
construction at $1.000$ retention and are not evidence of blackout tolerance. Its cells also cover 150
validation scenes rather than the full 6019-frame set. Its family ordering matches the main benchmark:
raindrop is its most damaging image degradation, at $0.13$ retention at S5 and already $0.34$ at S3,
which is worse at S3 than gaussian blur is at S5. A model trained to predict what the camera will see
next is, if anything, more exposed to local refractive distortion than one trained only to imitate
trajectories, which is the opposite of what a generative-prior argument would predict. Per-cell numbers
are in \texttt{results/epona\_summary.csv}.

\begin{table}[t]\centering\small
\caption{\textbf{Epona, an autoregressive world model, on the same ladder} (retention at $1$\,s, its native
horizon; $150$ scenes per cell). It shows the camera-reliant profile and reproduces the raindrop finding
independently: raindrop is its most damaging family at both severities, worse at S3 than gaussian blur
is at S5. Epona conditions on a single front camera, so blackout S3 darkens nothing it reads and its
$1.000$ is a protocol artifact rather than tolerance.}
\label{tab:epona}
\setlength{\tabcolsep}{4pt}
\begin{tabular*}{\linewidth}{@{\extracolsep{\fill}}lcc@{}}
\toprule
corruption & S3 & S5 \\
\midrule
gaussian blur      & 1.011 & 0.509 \\
jpeg               & 0.841 & 0.373 \\
motion blur        & 0.618 & 0.372 \\
\midrule
mud splash         & 0.881 & 0.457 \\
lens dirt          & 0.634 & 0.578 \\
occlusion          & 0.618 & 0.448 \\
raindrop           & \textbf{0.337} & \textbf{0.126} \\
\midrule
camera blackout    & 1.000 & 0.106 \\
\bottomrule
\end{tabular*}
\end{table}

\noindent\textbf{Epona, in full.} Its retention at $1$\,s is $0.51$ under gaussian blur S5, $0.37$ under
motion blur and JPEG S5, and $0.11$ with the front camera dark, which is the camera-reliant profile
rather than the ego-status-leaning one. Absolute meters are not comparable with the nuScenes planners
because its horizon is $1.5$\,s rather than $3$\,s; retention is, being a within-model ratio at a fixed
horizon.

\noindent\textbf{OpenDriveVLA.} For OpenDriveVLA~\cite{opendrivevla} we state the flat-on-both-axes result carefully because it concerns a
released model. Injection is verified three ways for every cell. The injector is exercised before
inference, the loader logs that it is live, and a cell whose predictions are byte-identical to the clean
cell aborts before scoring. The all-cameras-dark run differs from clean by construction and by checksum.
We evaluate the released $0.5$B variant and say so; larger variants may weight their cameras
differently. Placed in Table~\ref{tab:visrel} on the identical two axes, meaning the same nine quality families at S5
and the same blackout and dropout cells, it lands at $0.943$ and $0.964$, nearer the ego-only floor than
any of the six planners.

\section{Closed-loop detail}
\label{supp:carla}
\noindent\textbf{Route traces.} Most scored route-148 traces follow the turn, while incomplete traces
include a snow-S5 run circling near the start and a stalled blackout-S5 run
(Fig.~\ref{fig:carla_anchor}). That is how an anomalous trace can coexist with a high complete-case mean,
and it is the reason Sec.~\ref{sec:carla} reports identification bounds rather than the complete-case mean
alone.

\noindent\textbf{Why SimLingo absorbs image quality.} SimLingo follows a GPS route of target waypoints
with a PID controller over its ego speed and replans every frame, so the destination and proprioception
are not visual, and per-tick replanning can damp image-quality effects. The camera remains load-bearing:
with it fully removed but route and ego state intact, complete-case DS falls to $19.4$.

\noindent\textbf{Scoring rule.} Routes the simulator terminates because the agent stopped making
progress carry a real route-completion percentage and score, for example one blackout-S5 run at DS
$39.0$ and $88.8\%$ completion, and we average them rather than dropping them. Under a severe corruption
that termination is the outcome. Only infrastructure failures, meaning agent setup and simulator
crashes, which measure nothing, are excluded.

\noindent\textbf{TCP instability.} TCP's null-control movement is concentrated rather than spread. Its
clean score on route 180 is bimodal across seeds at $60/100/100/60$, and that route flips in every cell
including the null, while routes 168 and 196 move in one of eight null runs. TCP is also a weaker driver
than SimLingo on this slice, with a clean mean of $68.8$ against $90.0$, and it never completes route
157 at all, at DS $3.26$ and $0.49$, so we exclude that route as a floor case rather than average it. A
planner sitting near its decision boundaries produces run-to-run variance comparable to the corruption
effect, and a sample this size cannot resolve it.

Per-run scores for all three towns and both agents are released in
\texttt{results/carla/ds\_summary\_town03.csv}, \texttt{ds\_summary\_town05.csv}, and
\texttt{ds\_summary\_tcp.csv}. The paired comparison is \texttt{results/carla/tcp\_vs\_simlingo.py} and
the injection evidence is \texttt{results/carla/check\_injection.py}.

\section{Protocol for adding a planner}
\label{supp:protocol}
To add a planner, run its clean evaluator once, replay the same validation split under S1 to S5 for each
corruption family, and report both the clean score and the retention curve. Corruption is applied in the
loader, so no corrupted copies are written and the model's own evaluator is untouched.

We discourage collapsing the benchmark to one grand robustness number. The per-family knees are the
useful signal, because a method can improve blur tolerance while leaving camera loss or temporal dropout
unchanged, and a single average hides exactly that. When a model consumes route, ego-state, or other
non-camera inputs, the curve must be read together with the information-removal axis, so that a flat
curve is not mistaken for visual robustness.

Adding a planner requires one dataloader hook and no change to the benchmark. The injector is a single
dependency-light module with a stable entry point,
\texttt{corrupt\_one(family, image, severity, camera, frame, seed)}, and the corruption specification is
emitted from the same constants the injector uses, so a severity ladder cannot drift from its
documentation.

\noindent\textbf{Versioning and maintenance.} We will version the specification, keep the released
per-cell CSVs immutable per version, host the de-anonymized repository with issue tracking, and add
models and families as pull requests against a fixed ladder. Anything that changes an existing operator
becomes a new major version rather than a silent edit, so previously published numbers stay
reproducible.

\section{Full cross-corruption montage}
\label{supp:montage}
\begin{figure*}[t]\centering
  \includegraphics[width=\textwidth]{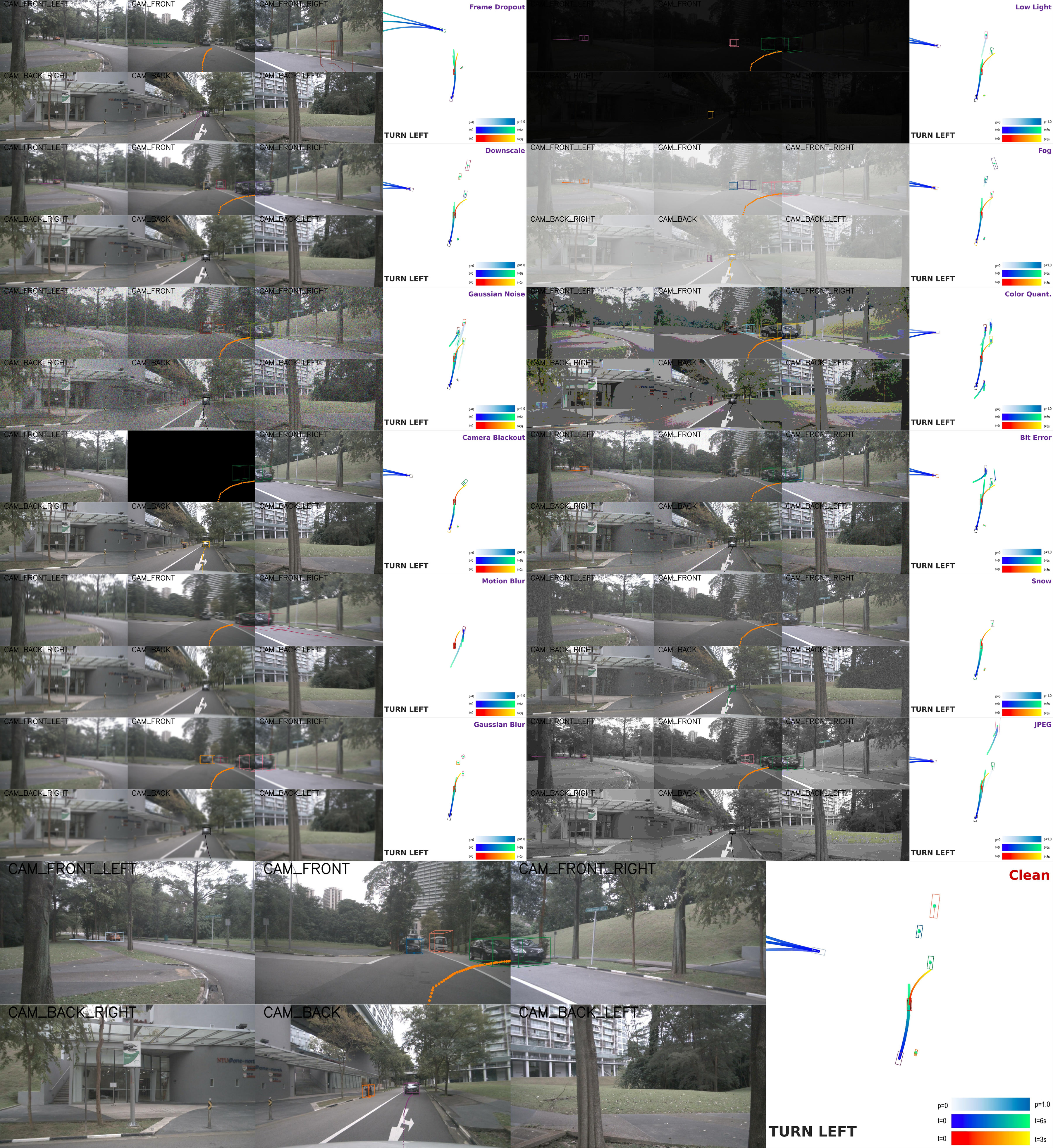}
  \caption{\textbf{UniAD cross-corruption qualitative comparison on one nuScenes scene.} Six
  surround-view images and the planned ego trajectory under each S5 corruption, clean for reference.}
  \label{supp:fig:montage}
\end{figure*}

Figure~\ref{supp:fig:montage} is the full version of Fig.~\ref{fig:comparison}. Each block shows all six
surround-view camera images and UniAD's planned trajectory under one S5 corruption, with the clean
scene for reference. The paper body carries a five-row crop because the camera tiles are unreadable at
column width.

\end{document}